\documentclass[%
 aip,
cp,  
 amsmath,amssymb,
 reprint,%
]{revtex4-2}

\usepackage{graphicx}
\usepackage{dcolumn}
\usepackage{bm}

\usepackage[utf8]{inputenc}
\usepackage[T1]{fontenc}
\usepackage{mathptmx} 
\usepackage{subcaption}
\usepackage{caption}
\usepackage{indentfirst}
\begin{document}

\title{Partial Coherence for Object Recognition and Depth Sensing}

\author{Zichen Xie} 
 \affiliation{
  School of Physics, Huazhong University of Science and Technology, Luoyu Road 1037, Wuhan, 430074, People’s Republic of China
}
\author{Ken Xingze Wang}%
 \email[Corresponding author: ]{wxz@hust.edu.com}
 \affiliation{
  School of Physics, Huazhong University of Science and Technology, Luoyu Road 1037, Wuhan, 430074, People’s Republic of China
}

\date{\today} 

\begin{abstract}
We show a monotonic relationship between performances of various computer vision tasks versus degrees of coherence of illumination. We simulate partially coherent illumination using computational methods, propagate the lightwave to form images, and subsequently employ a deep neural network to perform object recognition and depth sensing tasks. In each controlled experiment, we discover that, increased coherent length leads to improved image entropy, as well as enhanced object recognition and depth sensing performance.
\end{abstract}

\maketitle


Over the past few years, extensive research has been conducted on the use of ambient light for passive illumination or controlled active illumination with additional light sources~\cite{ikeuchi2020active}. It has been established that active illumination, such as structured light, can achieve specific applications in certain scenarios compared to passive illumination~\cite{batarseh2018passive,forbes2021structured,scharstein2003high}. For instance, coherent illumination can achieve non-line-of-sight recognition in specific scenarios~\cite{Lei_2019_CVPR,Ando:15,Klein2016}. This is because coherent illumination provides more information for recognition, owing to its higher signal-to-noise ratio and possible inclusion of phase information~\cite{796348,doi:10.1080/713818245}, compared to passive illumination using natural light. Even in general passive lighting scenarios, replacing natural light with coherent light to illuminate objects should provide more information. Therefore, in theory, the use of coherent light sources can enhance computer vision tasks.

Although coherent light offers numerous advantages, non-ideal light sources are inevitably affected by environmental perturbations and have a certain spatial scale, spectral width. Consequently, in practical scenarios, the emitted light from the vast majority of sources is not perfectly coherent. Therefore, it is necessary to study the impact of light source coherence on illumination. However, although there has been considerable research on coherence, no study has yet explored the relationship between coherence, recognition accuracy, and image information. The contribution of this study is to reveal, for the first time, the relationship between the coherence of a light source and object recognition and depth sensing accuracy, through controlled experiments. Multiple application scenarios were established to perform object recognition and depth sensing in both common direct imaging scenes and complex imaging scenes where interference patterns of scattered light were obtained. Under different scenarios, datasets and tasks, all results indicate that coherence improves information entropy and thus recognition and sensing.

\begin{figure}[htb]
	\includegraphics[width=0.95\linewidth]{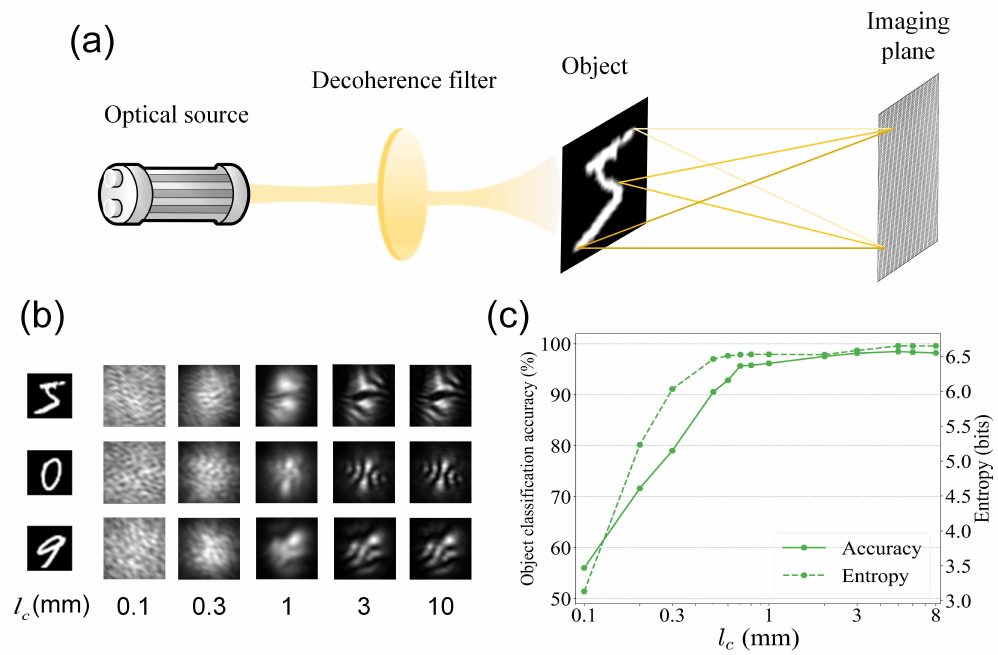}
	\caption{\label{fig:1}
		Direct imaging scenario experimental setup and results.
		\textbf{(a)} Schematic diagram of basic scene setup. The coherent light emitted by the source becomes partially coherent after decoherence. 
            \textbf{(b)} Example images of imaging results of different objects under varying degrees of coherence. The images with a low $l_{c}$ value are also quite dark, so we have normalized the images to improve their visibility.
		\textbf{(c)} Object recognition accuracy and two-dimensional information entropy curves.} 
\end{figure}

The direct imaging scenario is depicted in Figure~\ref{fig:1}(a). We refer to this scenario as the direct imaging scenario.
In this scenario, a light source illuminates the object, which can be viewed as a spatial light modulator (SLM) loaded with an image. The SLM serves to modify the transmittance of each pixel, with higher gray values corresponding to lower pixel transmittance. The scattered light from the object propagates and is captured at the imaging plane. To obtain imaging results under different coherence degrees, the coherence of the light source is adjusted, and a deep neural network is utilized to identify the results.
 
The light source used in this scenario is a coherent light source with a wavelength of 635 nm, and we assume that the light wave initially emits is in the plane wave form. The light wave emitted from the coherent light source experiences instantaneous coherent decay, becomes partially coherent light. This process can be regarded as acquiring a new partially coherent light source. Henceforth, all references to light sources in the text will pertain to this newly acquired partially coherent light source.
The object size is set at 6 mm x 6 mm, with a pixel size of 512 x 512, and the distance from the object to the imaging plane is set at 2.5 meters.
The propagation process of the light field is accomplished by using the Fast Fourier Transform (FFT) in the form of the angular spectrum~\cite{goodman2005introduction,press2007numerical,mandel1995optical}. 

The 2D objects illuminated by partially coherent light in Figure~\ref{fig:1}(a) are from the MNIST dataset~\cite{lecun1998mnist,lecun1998gradient}, which is one of the most commonly used datasets in machine learning. It contains images of 10 digits from 0 to 9 written by different people. We choose 5000 of these images as our object set and obtain the imaging results of different objects under illumination of light sources with different coherence levels, as shown in Figure~\ref{fig:1}(b). To improve the clarity of the images for the reader, we normalize them so that images with lower $l_{c}$values are brighter than their actual brightness. However, it is important to note that images with low $l_{c}$values are naturally quite dark.

Figure~\ref{fig:1}(b) demonstrates the significant impact of the coherence of the light source on imaging results, which in turn affects the accuracy of object recognition. The accuracy curve shown in Figure~\ref{fig:1}(c) quantitatively reflects this relationship and is the key result of the study, with the x-coordinate representing the transverse coherence length of the light source~\cite{voelz2011computational}, denoted by $l_{c}$. A higher value of $l_{c}$ indicates better coherence of the light source, while the y-coordinate represents the accuracy of object recognition. The variables in the experiment were tightly controlled, and the sole variation among each data point in the figure is the $l_{c}$ variable. The graph illustrates that an improvement in the coherence of the light source results in an increase in the accuracy of object recognition. This trend is particularly evident when the coherence of the light source is low, but when $l_{c}$reaches 1mm, the improvement in recognition accuracy is small, and when $l_{c}$is raised from 1mm to 8mm, the recognition accuracy is improved by less than 2\%. This shows that the actual difference from using these two light sources with very different coherence degrees is small. 

In real-world scenarios, the presence of ambient light and thermal disturbances poses challenges in achieving perfect coherence. Researchers aiming to minimize the impact of incoherent light often resort to lasers and spectral filters. However, as illustrated by the results presented in Figure~\ref{fig:1}(c), we observe that as long as the light source maintains a reasonable level of coherence, the overall recognition accuracy remains largely unaffected. This suggests that complete coherence may not be essential in many practical scenarios, and satisfactory results can be achieved without the requirement for a fully coherent light source.

One question is what elements of the image are affected by the coherence of the light source that change the image quality. From the example shown in Figure~\ref{fig:1}(b), we can observe that the coherence performance of the light source has an impact on the image pattern. Specifically, the image pattern at the $l_{c}$ of 0.1 mm is entirely distinct from that at the $l_{c}$ of 10 mm. Nevertheless, we contend that the more fundamental alteration is the quantity of information encapsulated within the image, and the shift in the image pattern merely serves as an intuitive manifestation of the change in the amount of information. We contend that the coherence of the light source plays a crucial role in determining the quantity of information that is present in the resulting image, which, in turn, affects the accuracy of image recognition.

To substantiate this perspective, we introduce the concept of two-dimensional information entropy~\cite{abutaleb1989automatic,larkin2016reflections} as a measure of an image's information content. Unlike one-dimensional information entropy, the two-dimensional information entropy calculates the entropy of an image based on the probability distribution of intensity values within a local neighborhood around each pixel. This approach accounts for the spatial relationships between pixels and can identify regions of the image that contain more complex or diverse information. For instance, in an image with half black and half white pixels, the left half of the image could be entirely black while the right half is entirely white, or the black and white pixels could be spread out evenly, or the distribution of black and white pixels could be uneven. In all three cases, the one-dimensional information entropy values would be identical, despite the obvious difference in the amount of information conveyed. This issue is resolved by using two-dimensional image entropy.

Two-dimensional information entropy is calculated as follows: We divide the grayscale into 256 levels (0-255). At each pixel, we calculate the average gray value of the neighborhood. The gray value i of the pixel and the average value j of the neighborhood form a pair (i, j). The total number of occurrences (frequency), f(i, j), of a pair (i, j) divided by the total number of pixels, $N^2$, defines the probability of this pair, denoted as $P_{ij}$, i.e.
\begin{equation}
	\label{eq1}
	P_{i j}=f(i, j) / N^2
\end{equation}
Similar to one-dimensional entropy, two-dimensional entropy, H, can be calculated as follows:
\begin{equation}
	\label{eq2}
	H=-\sum_{i=0}^{255} \sum_{j=0}^{255} P_{i j} \log _2 P_{i j}
\end{equation}

We computed the average two-dimensional information entropy values for images corresponding to different coherence degrees, and the results are presented in the entropy curve depicted in Figure~\ref{fig:1}(c). The results show that with the increase of the coherence degree of two-dimensional information entropy increases firstly and then tends to saturation, the amount of imaging information is also this trend, which indicates that the increase of the coherence degree when the low coherence degree can improve the amount of information to a large extent, and when the coherence degree reaches a certain level, and continue to increase the degree of coherence, even if it will lead to large changes in the imaging results, but the effective amount of information in the imaging results has not been changed greatly, and the recognition accuracy is also therefore not a big change.
Importantly, although the physical quantities represented by the two-dimensional entropy curve and the precision curve in Figure~\ref{fig:1}(c) are distinct, they exhibit remarkably similar trends, which responds to the validity of the two-dimensional information entropy as a measure of the quality of the imaging process.


Here we introduce the decoherence filter shown in Figure~\ref{fig:1}(a) to describe our method of simulating light sources with varying coherence levels.
As it is straightforward to simulate a fully coherent source, we simulate a source with reduced coherence by reducing the light field correlation. 
A commonly employed method for generating partially coherent light in laboratory settings involves the utilization of SLM to implement random phase screens. This technique entails generating random phase modes on the SLM, whereby the phase of the incident beam is modulated based on the loaded random phase modes. To simulate this process computationally, we employ a dynamic random phase screen approach. Specifically, We place a dynamic random complex phase screen along a completely coherent light field propagation path~\cite{voelz2011computational,morgan2010measurement}, perform multiple iterations of random phase screen generation and beam propagation, and simulate the dynamic changes of the random phase screen over time. The resulting light intensity pattern is obtained by averaging the output of each iteration. The autocorrelation function of the optical field after passing through the random phase screen can be expressed as follows:

\begin{equation}
    \label{eq1}
	R\left(x^{\prime}, y^{\prime}\right)=\exp \left(-\frac{x^{\prime 2}+y^{\prime 2}}{l_{c}^2}\right)
\end{equation}

\begin{figure}[htb]
	\centering
	\includegraphics[width=0.85\linewidth]{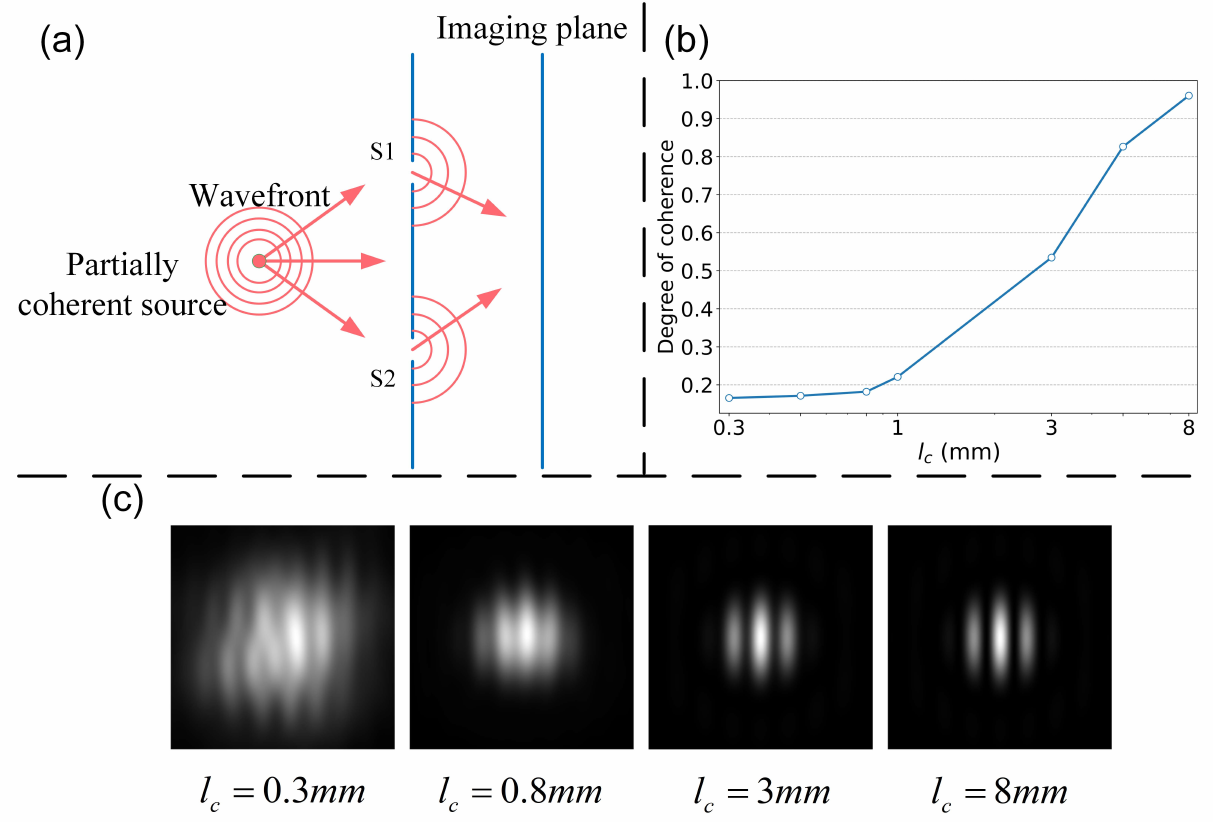}
	\caption{\label{fig:2}
            The degree of coherence measurement schematic and results.
		\textbf{(a)} Schematic diagram of measurement optical path.
            \textbf{(b)} Quantitative relationship between $l_{c}$ and degree of coherence.
            \textbf{(c)} Interference fringe images at different degrees of coherence.
	}
\end{figure}

The coherence degree can be regulated by adjusting the transverse coherence length, represented by $l_{c}$.
As indicated by Equation (3) , a larger value of $l_{c}$ corresponds to improved coherence. 
As $l_{c}$ approaches infinity, the transmitted beam remains fully coherent, whereas when $l_{c}$ approaches zero, the transmitted beam is considered incoherent. In order to better establish the quantitative relationship between $l_{c}$ and coherence, we use the classical Young's double hole interference to measure the degree of coherence between two points on the optical field~\cite{zernike1938concept}, and the measurement optical path is set up as shown in Figure~\ref{fig:2}(a), in which the hole diameter is 1 mm, the two-hole spacing is 5 mm, and the distance from the hole to the imaging plane is 25 mm. Figure~\ref{fig:2}(c) shows the interference fringes when $l_{c}$ is equal to 0.3 mm, 0.8 mm, 3 mm, 8 mm. It can be seen that the contrast and sharpness of the interference fringes are decreasing as $l_{c}$ decreases. Figure~\ref{fig:2}(b) illustrates the monotonic quantitative relationship between $l_{c}$ and the degree of coherence, and thus, we demonstrate that $l_{c}$ is a characterisation and measure of coherence.

\begin{figure}[htbp]
	\includegraphics[width=0.95\linewidth]{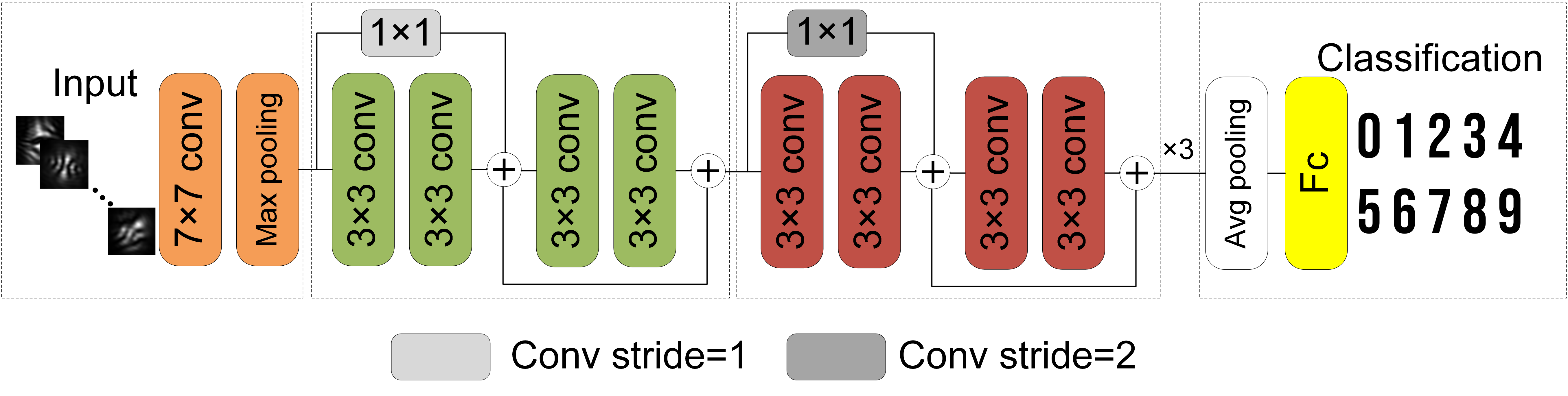}
	\caption{\label{fig:3}
		The neural network architecture utilized for the image classification.
		It consists of a combination of convolutional and fully connected layers.
		The abbreviation "Conv" stands for a convolutional layer, "Avg" represents an average pooling layer, and "Fc" denotes a fully connected layer.
	}
\end{figure}

Although one would find it difficult to extract information from the images obtained above, it is perfectly possible for deep neural networks to extract information about objects in different images. Thus we can train a neural network to test it, as we did for a common computer vision problem.Resnet is widely used for such problems and due to its unique residual block design~\cite{he2016deep,2016residual2}, it is easy to train a deep neural network on it. The architecture of the ResNet-18 model is depicted in Figure~\ref{fig:3}. We randomly split 90\% of the 5000 images obtained under each $l_{c}$ as a training dataset and the remaining 10\% as a test dataset, which are fed into Resnet. It is noteworthy that, in order to eliminate the potential interference,  we fed the images directly into the recognition program without any modification, and we refrained from applying any pre-processing techniques, such as normalization, on the imaging results.

\begin{figure}[htbp]

	\includegraphics[width=1\linewidth]{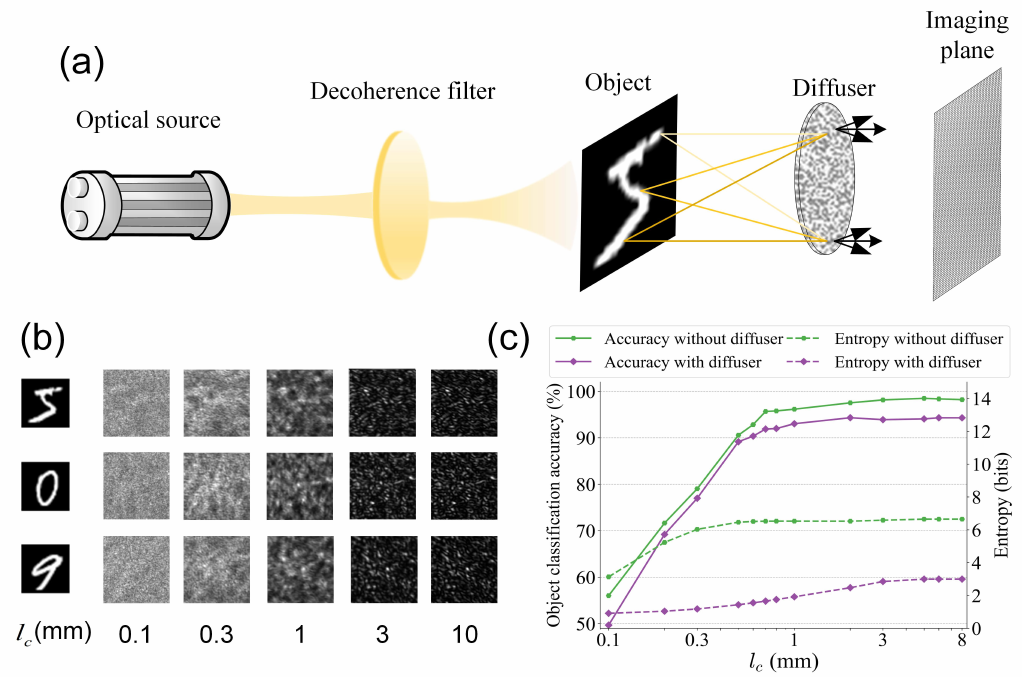}
	\caption{\label{fig:4}
            Scattering imaging scenario experimental setup and results.
		\textbf{(a)} Schematic diagram of the scenario after the addition of a diffuser.
            \textbf{(b)} Image examples of imaging results of different objects with different coherence degrees.
            \textbf{(c)} Accuracy curves and two-dimensional information entropy curves with and without diffuser.
	}
	
\end{figure}

To further substantiate the universality of our findings, supplementary experiments were conducted in more complex scenario. 
As illustrated in Figure~\ref{fig:4}(a), a diffuser (ground glass) was introduced into the optical path to create a more intricate scene. 
This entailed adding the ground glass to the middle of the optical path, thus inducing scattering of the light from the object, which resulted in the formation of an interference pattern at the imaging plane~\cite{tan2019imaging}.

Similar to the direct imaging scenario, we used the MNIST dataset as the object set to obtain scattering imaging results for different objects illuminated by light sources with different coherence degrees. Figure~\ref{fig:4}(b) presents the results of the simulation, which exhibit that the imaging outcomes, in the presence of increased ground glass scattering, appear as speckle patterns~\cite{katz2014non,franccon2012laser} and that these patterns change as the degree of coherence changes. 
 	

The speckle images were inputted into the ResNet-18 model. Although it is difficult for the human eye to discern the information contained in the speckle image, the deep learning network can extract information for recognition, as shown in Figure~\ref{fig:4}(c). A comparison of the results with those from the simple scenario (Figure~\ref{fig:1}(a)) demonstrates that, although the accuracy in the scattering scene decreased, the overall trend remains. Additionally, we calculated the two-dimensional information entropy in the scattering scene and the results are displayed in Figure~\ref{fig:4}(c) (entropy with diffuser) . It can be observed that the effect of coherence on the entropy value in the scattering scenario is similar to that in the direct scenario.
	

However, one thing to note is that the accuracy with the addition of diffuser does not drop much compared to the accuracy without diffuser, but the entropy is much smaller. Our explanation for this is that speckle patterns are highly correlated and repetitive, resulting in a limited range of pixel values~\cite{castleman1996digital}. Speckle patterns are generated by the interference of coherent light waves, which produce a random pattern of bright and dark spots that are highly correlated with each other. As a result, speckle images have high contrast, the distribution of pixel values in the speckle image tends to peak values with relatively few distinct~\cite{goodman1975statistical,briers2013laser}. Speckle patterns are much less random and complex than other types of images, such as natural scenes or photographs. This means that there is less information content in the image, which is reflected in lower entropy values. However, it should be noted that the low entropy value of the speckle image does not mean that it contains less effective information, and the speckle image still contains a lot of effective information~\cite{briers1996laser,leger1975optical,BARADIT2020106009,yang1995precision}.

\begin{figure}[htbp]
	\includegraphics[width=0.95\linewidth]{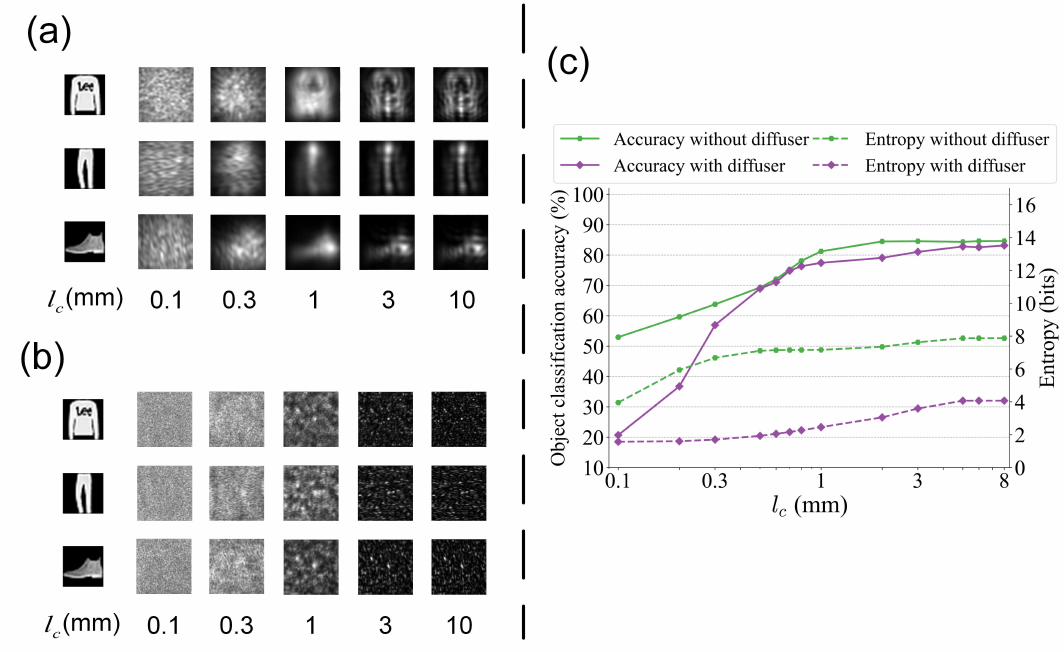}
	\caption{\label{fig:5}
		Examples and results of imaging of Fashion-MNIST object set at different coherence levels.
		\textbf{(a)} Example of imaging for the scenario depicted in Figure ~\ref{fig:1}(a).
		\textbf{(b)} Example of imaging for the scenario depicted in Figure ~\ref{fig:4}(a).
            \textbf{(c)} Accuracy curves and two-dimensional information entropy curves with and without diffuser.
	}
	
\end{figure}

In addition to the MNIST dataset, a more complex dataset, the Fashion-MNIST dataset~\cite{xiao2017fashion}, was also utilized as the object set in our object recognition experiments, with the aim of demonstrating that object complexity does not affect our conclusions.
Examples of simulation results are presented in Figure~\ref{fig:5}(a) and Figure~\ref{fig:5}(b). 

The recognition results are presented in the accuracy curve of Figure~\ref{fig:5}(c). As evident from the figure, the recognition task becomes more challenging, and the recognition accuracy decreases with an increasing complexity of the object set. Specifically, when the scattering imaging scenario employs the Fashion-MNIST dataset as the object set, the recognition accuracy is only around 20$\%$ under low coherence light source illumination. However, with an increase in the coherence of the light source, the trend of increasing accuracy persists, and with sufficient coherence, the accuracy for these cases eventually surpasses 80$\%$. Furthermore, we observe that changing the dataset does not affect the conclusion that coherence improves the informativeness of the imaging results, as shown by the entropy curve in Figure~\ref{fig:5}(c). 
	
\begin{figure}[htp]

	\includegraphics[width=0.95\linewidth]{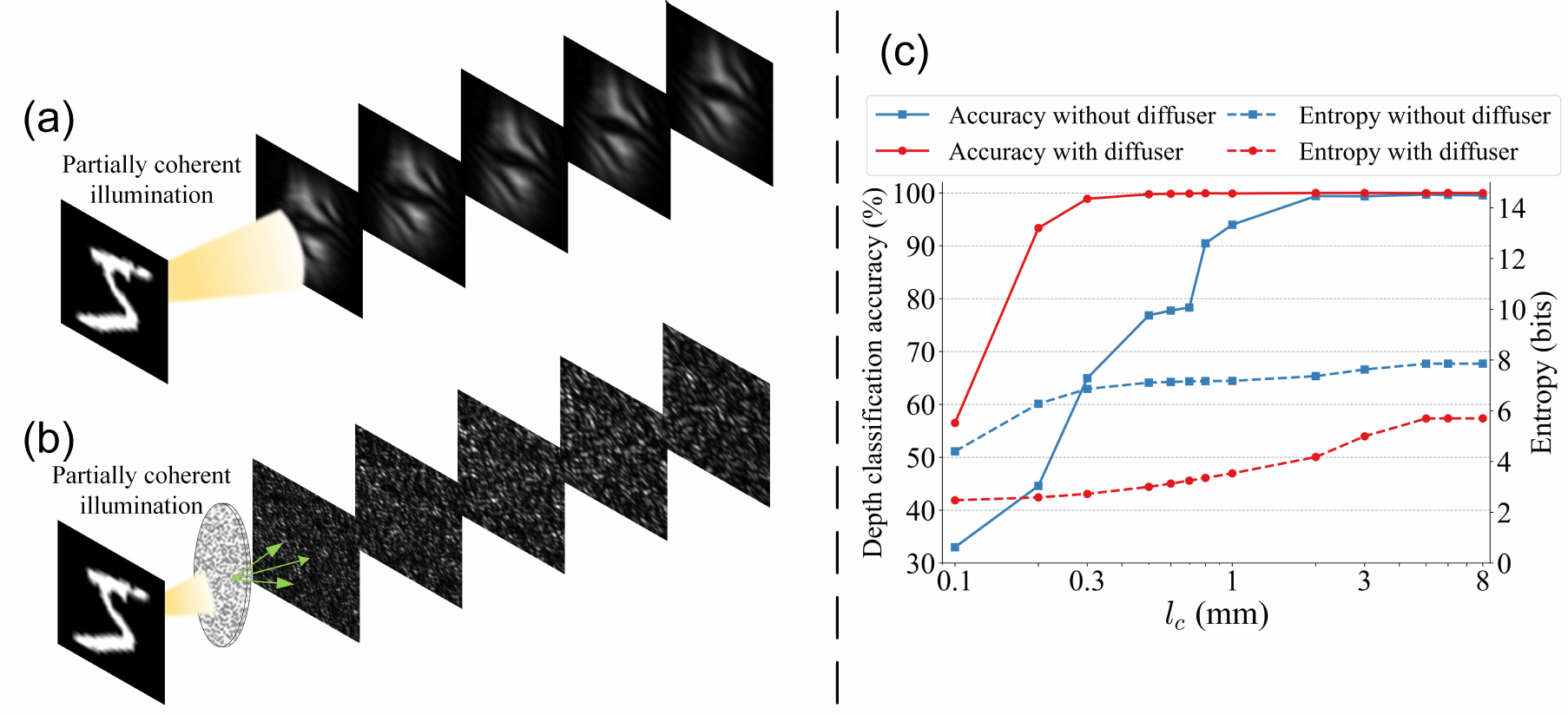}
	\caption{\label{fig:6}
		depth sensing schematics and results.
		\textbf{(a)} depth sensing for direct imaging scenario.
		\textbf{(b)} depth sensing of scattering scenario after adding ground glass.
  		\textbf{(c)} Accuracy curves and two-dimensional information entropy curves for depth sensing.
	}
	
\end{figure}

To illustrate that our conclusions do not apply only to specific tasks, we established two depth sensing scenarios. depth sensing  refers to the ability of a machine or computer to perceive and understand the spatial depth of objects within an image or scene~\cite{torralba2002depth}. The setup for the depth sensing scenarios closely follows that depicted in Figure~\ref{fig:1}(a) and Figure~\ref{fig:4}(a), with the exception that the imaging plane is placed at different distances. As shown in Figure~\ref{fig:6}(a) and Figure~\ref{fig:6}(b), the imaging plane was positioned at five different depths, resulting in varying images due to changes in the distance between the object and the imaging plane.

A total of 5000 randomly selected images from the Fashion-MNIST dataset were used as objects, and images were captured at different depths. The results, presented in Figure~\ref{fig:6}(b), demonstrate that increasing the coherence of the light source improves depth sensing accuracy.

Interestingly, it was observed that the accuracy of complex scenes with ground glass scattering was consistently higher than that of simple scenes. The accuracy rapidly increased with the coherence of the light source, reaching maximum saturation accuracy quickly. This phenomenon can be explained by the differences in speckle patterns scattered by ground glass at different depths. The deeper the depth, the larger the speckle size and the fewer the number of speckles in a given unit area~\cite{hu2020does,funamizu2007generation,crammond2013speckle}. Larger speckles provide more texture to the pattern and, therefore, more spatial information in the image~\cite{crammond2013speckle}. In fact, 
as evidenced in Figure~\ref{fig:6}(b), the average speckle size exhibits a linear relationship with distance~\cite{hu2020does}. This feature of speckle images is so distinctive that a deep neural network can easily determine the depth of an image.


	

In summary, this study confirms that light source coherence has a positive impact on various computer vision tasks. This is due to the fact that a light source with high coherence can increase the amount of image information, which is reflected in the change in information entropy. Our study has demonstrated the robustness and interpretability of our conclusions.

However, the experimental scene settings are relatively ideal, and real-world scenarios are typically more complex. Thus, further research can be conducted on more complex scenes, such as optical coherence tomography in biomedical imaging, autonomous driving, and other realistic scenarios.

\section{data availability}
The data utilized in this study are openly available in the MNIST handwritten digit database and the Fashion-mnist database, accessible at http://yann.lecun.com/exdb/mnist/ and https://github.com/zalandoresearch/fashion-mnist, respectively~\cite{lecun1998mnist,xiao2017fashion}.

\nocite{*}
\bibliography{aipsamp}

\end{document}